\pdfoutput=1

\documentclass[11pt]{article}
\usepackage[final]{acl}
\usepackage{times}
\usepackage{latexsym}
\usepackage[T1]{fontenc}
\usepackage[utf8]{inputenc}
\usepackage{microtype}
\usepackage{booktabs}
\usepackage{inconsolata}
\usepackage{graphicx}
\usepackage{amsmath}
\usepackage{amssymb} 
\usepackage{subcaption}
\usepackage{multirow}
\title{GTA: Supervised-Guided Reinforcement Learning for Text Classification with Large Language Models}

\author{
 \textbf{Min Zeng},
 \textbf{Jingfei Sun},
 \textbf{Xueyou Luo},
 \textbf{Caiquan Liu},
 \textbf{Shiqi Zhang},
 \textbf{Li Xie}, 
 \textbf{Xiaoxin Chen}
\\
vivo AI Lab
\\
 \small{
  \href{mailto:zengmin.ai@vivo.com}{zengmin.ai@vivo.com}
 }
}

\begin{document}
\maketitle
\begin{abstract}
In natural language processing tasks, pure reinforcement learning (RL) fine-tuning methods often suffer from inefficient exploration and slow convergence; while supervised fine-tuning (SFT) methods, although efficient in training, have limited performance ceiling and less solid theoretical foundation compared to RL. To address efficiency-capability trade-off, we propose the \textbf{G}uess-\textbf{T}hink-\textbf{A}nswer (\textbf{GTA}) framework that combines the efficiency of SFT with the capability gains of RL in a unified training paradigm. GTA works by having the model first produce a provisional guess (optimized via cross-entropy loss), then reflect on this guess before generating the final answer, with RL rewards shaping both the final output and the format of the entire GTA structure. This hybrid approach achieves both faster convergence than pure RL and higher performance ceiling than pure SFT. To mitigate gradient conflicts between the two training signals, we employ loss masking and gradient constraints. Empirical results on four text classification benchmarks demonstrate that GTA substantially accelerates convergence while outperforming both standalone SFT and RL baselines.
\end{abstract}

\section{Introduction}
Text classification, as a foundational task in natural language processing (NLP), has been widely employed for sentiment analysis \citep{sentiment}, intent recognition \citep{intent}, and news categorization \citep{news}. Early NLP solutions primarily relied on rule-based systems and statistical models—including hidden Markov models (HMMs) and support vector machines (SVMs)—to learn patterns from annotated corpora \citep{svm}. The emergence of deep learning and Transformer \cite{Transformers}  architectures dramatically enhanced classification performance, with Bidirectional Encoder Representations from Transformers (BERT)’s \citep{bert} bidirectional pre-training paradigm capturing rich contextual representations and achieving breakthroughs across multiple tasks. 

Large language models (LLMs)—such as GPT \citep{gpt4}, Llama \citep{llama}, Qwen \citep{qwen}, and DeepSeek \citep{deepseek}—have demonstrated remarkable capabilities across numerous NLP tasks, including text classification \citep{textclass}. While SFT has been the predominant approach to adapt these models for specific tasks, it faces inherent limitations: SFT methods directly learn to produce correct answers without explicit reasoning, leading to limited generalization capabilities and performance ceilings. Chain-of-thought (CoT) prompting \citep{Cot}—a technique that guides models to generate intermediate reasoning steps before producing final answers—has shown significant improvements across various reasoning tasks \citep{kojima2022large, BCOT}. However, applying CoT within the SFT paradigm requires extensive human annotation of reasoning chains, resulting in substantial costs and susceptibility to annotator biases and quality inconsistencies \citep{tan2024large, byun2024ares}.

RL offers a promising alternative that can theoretically overcome these limitations by combining the benefits of CoT reasoning with optimization-based learning \citep{xu2025towards, wang2024reinforcement}. From reinforcement learning from human feedback (RLHF) \citep{rlhf} to advanced frameworks like group relative policy optimization (GRPO) \citep{grpo}, RL techniques can explore and optimize intermediate reasoning processes without requiring manually annotated reasoning chains. This approach holds particular promise for enhancing model performance beyond what SFT can achieve. However, the application of RL to text classification tasks remains challenging due to fundamental limitations: unlike supervised learning's direct approach, RL methods must discover optimal reasoning strategies through self-guided exploration—a process often hampered by inefficient exploration, slow convergence, and potential training instability \citep{chen2025research}. These efficiency challenges have hindered pure RL-based methods from consistently outperforming SFT approaches despite their stronger theoretical foundation. 

To address these challenges, this work introduces a novel Guess–Think–Answer (GTA) framework that seamlessly integrates the advantages of SFT and RL within a unified single-stage training process. In our approach, the model first generates an intuitive guessed answer, then engages in a "think" step—reasoning explicitly over the guessed answer and the input question—before producing a refined final answer. 
Our main contributions can be summarized as follows:
\begin{enumerate}
\item We propose a novel GTA framework that structures the reasoning process into three distinct phases: an initial intuitive guess, an explicit reasoning step that reflects on this preliminary prediction, and a refined final answer that incorporates this reasoning.

\item We develop a unified training approach that seamlessly integrates SFT and RL within a single-stage process. Our method applies cross-entropy loss to the guessed answer while optimizing the reasoning process and final answer through RL-based rewards. To ensure effective cooperation between these learning paradigms, we introduce a specialized loss masking strategy and gradient cosine adjustment technique that mitigates potential gradient conflicts.

\item Our framework eliminates the need for manual annotation of reasoning chains by enabling the model to spontaneously learn effective reasoning patterns through reinforcement. Experimental results demonstrate that GTA significantly outperforms both standard SFT baselines and state-of-the-art RL methods across multiple text classification benchmarks.
\end{enumerate}

\section{Related Work}
CoT prompting has emerged as a powerful technique to enhance the reasoning capabilities of LLMs by guiding them to generate intermediate reasoning steps before producing final answers \citep{Cot, kojima2022large, BCOT}. While CoT significantly improves performance across various reasoning tasks, its application in text classification often requires extensive human annotation of reasoning chains, leading to substantial costs and quality inconsistencies \citep{tan2024large, byun2024ares}. To address these limitations, researchers have explored RL approaches that can optimize model behavior without requiring manual annotation of intermediate steps \citep{xu2025towards, wang2024reinforcement}. 

Traditional RL methods in NLP, however, often suffer from inefficient exploration and slow convergence, making it challenging for pure RL-based methods to consistently outperform SFT approaches despite their stronger theoretical foundation \citep{chen2025research}. Recent advancements like the GRPO algorithm \citep{grpo} have improved RL efficiency by estimating baselines from group scores, thereby reducing computational costs. Additionally, when combining multiple learning objectives—such as SFT and RL—gradient conflicts can arise, leading to suboptimal convergence. Techniques such as gradient masking and cosine similarity adjustments have been developed to mitigate these conflicts by aligning gradients toward compatible directions \citep{pcgrad, chen2018gradnorm}, which inspires our approach to harmonizing supervised and RL signals within our proposed GTA framework.

\begin{figure*}[htbp]
	\centerline{\includegraphics[width=0.97\textwidth]{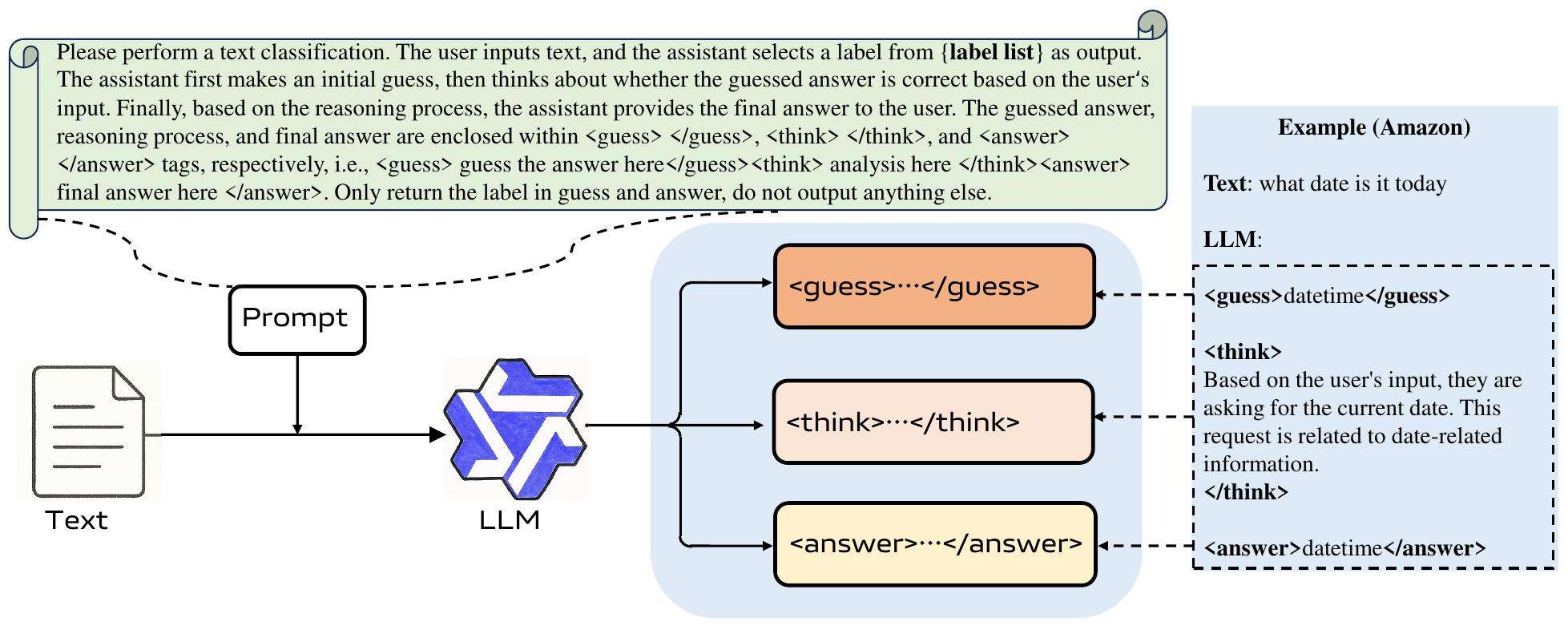}}
	\caption{Overview of the Guess–Think–Answer framework.}
	\label{gta}
\end{figure*}

\section{Methodology}
In this section, we present the proposed GTA framework in detail. We begin with an overview of the overall design, followed by the description of each component and training objective, highlighting how the method effectively integrates reinforcement learning with reasoning-oriented supervision.

\subsection{Prompt Design}

To accelerate convergence during RL training, we propose the GTA, which introduces a novel \texttt{Guess} stage to the conventional reasoning process. As illustrated in Figure~\ref{gta}, our prompt design guides the model to sequentially produce three components: \texttt{Guess}, \texttt{Think}, and \texttt{Answer}. The right side of the figure presents an example of a model-generated response adhering to this structure.

\begin{itemize}
	\item \textbf{Guess}: In this initial stage, the model generates a preliminary answer based on intuition or prior knowledge. This guess serves as a reference point for subsequent reasoning and the final answer. During training, the \texttt{Guess} component is supervised using cross-entropy loss.
	
	\item \textbf{Think}: Building upon the initial guess and the input question, the model produces a sequence of reasoning steps. This process aids in task comprehension and enhances the model's generalization capabilities.
	
	\item \textbf{Answer}: The final answer is generated by integrating insights from both the \texttt{Guess} and \texttt{Think} stages. This answer may align with or differ from the initial guess. The quality of the \texttt{Answer}, along with the overall output structure, is evaluated using a reward signal, which guides the RL component of the training.
\end{itemize}

By incorporating supervised signals in the \texttt{Guess} stage, our framework accelerates RL convergence and fosters the generation of interpretable reasoning processes.

\begin{figure*}[htbp]
	\centerline{\includegraphics[width=0.95\textwidth]{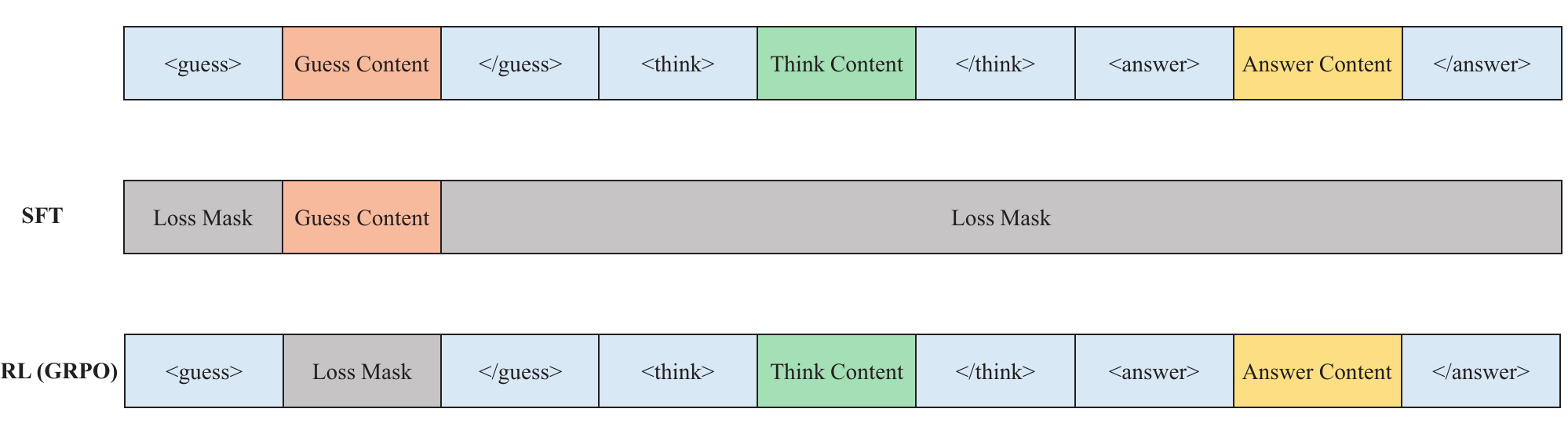}}
	\caption{Illustration of the loss masking strategy applied during training.}
	\label{lossmask}
\end{figure*}

\subsection{Training Objective}

 We propose a unified training framework that combines SFT and RL within a single optimization process. As shown in Figure~\ref{lossmask}, our approach exploits the GTA output format to assign distinct training objectives to the model’s various outputs.

\paragraph{SFT Loss.} For the \textit{Guess} segment, which represents the model's initial prediction based on intuition or prior knowledge, we employ a standard cross-entropy loss. To ensure that the loss computation focuses solely on this segment, we apply a masking strategy that assigns a special token (e.g., \texttt{-100}) to tokens outside the \textit{Guess} span. Formally, the SFT loss is defined as:
\begin{equation}
\mathcal{L}_{\text{SFT}} = -\sum_{t \in \mathcal{G}} \log P_{\theta}(y_t | y_{<t}, x)
\end{equation}
where $\mathcal{G}$ denotes the set of token positions corresponding to the \textit{Guess} segment, $y_t$ is the target token at position $t$, $y_{<t}$ represents the sequence of preceding tokens, and $x$ is the input text.

\paragraph{RL objective function.} 
In optimizing the model’s final output, we introduce the GRPO \citep{grpo} algorithm with minor modifications. GRPO improves sample efficiency by generating multiple candidate outputs for the same input prompt and computing relative advantages without a separate value function, thereby reducing training resource consumption. In LLMs, reward signals can be categorized into model-based and rule-based rewards; text classification tasks are particularly amenable to rule-based rewards, which are assigned by directly comparing the model’s final prediction to the ground-truth label. The reward definitions are as follows:
\begin{align}
	R_{\mathrm{format}} &=
	\begin{cases}
		1, & \text{if format correct},\\
		0, & \text{otherwise},
	\end{cases} \\[1ex]
	R_{\mathrm{accuracy}} &=
	\begin{cases}
		1, & \text{if classification correct},\\
		0, & \text{otherwise},
	\end{cases} \\[1ex]
	R_{\mathrm{total}}  &= R_{\mathrm{format}} + R_{\mathrm{accuracy}}.
\end{align}
where $R_{\mathrm{format}}$ denotes the format reward, which is granted whenever the model’s output adheres to the prescribed GTA format. $R_{\mathrm{accuracy}}$ denotes the accuracy reward, which is awarded when the model’s final prediction matches the true label. $R_{\mathrm{total}}$ represents the overall reward signal. The overall RL training objective is defined as follows:

\begin{align}
 	\mathcal{J}(\theta) =\ &
	\mathbb{E}\bigl[q \sim P(Q),\ \{o_i\}_{i=1}^G \sim \pi_{\theta_{\text{old}}}(\cdot|q)\bigr] \Bigg[ \frac{1}{G} \nonumber \\  &
	 \sum_{i=1}^G \frac{1}{|o_i|}\sum_{t\notin\mathcal{G}}  
	\,\min \Bigg(r_i(\theta), \nonumber \mathrm{clip} (
	r_i(\theta), \\&
	1 - \epsilon,\ 1 + \epsilon )
	\Bigg) 
	\hat A_{i,t} \Bigg] -\beta\,\mathbb{D}_{\mathrm{KL}}\bigl[\pi_\theta\Vert\pi_{\mathrm{ref}}\bigr]
\end{align}
where $G$ denotes the group size, representing the number of output sequences sampled in parallel for the same prompt. The $i$-th output sequence is denoted as $o_i$, where a loss mask is applied to the text within the \textit{Guess} segment to selectively include tokens in the loss computation. $\hat A_{i,t}=(R_{ \mathrm{total},i} - \mu ) / \sigma$ denotes the advantage function, obtained by subtracting the group’s mean reward $\mu$ from each individual reward and then dividing by the group’s reward standard deviation $\sigma$. The ratio $r_i(\theta) = \frac{\pi_\theta(o_i\mid q)}{\pi_{\theta_{\text{old}}}(o_i\mid q)}$ corresponds to the probability ratio between the new and old policies. The hyperparameter $\epsilon$ defines the clipping range for gradient updates, and $\beta$ adjusts the weight of the Kullback–Leibler (KL) divergence term. Since our training involves backpropagating two distinct loss components, we jointly constrain the magnitude of model updates using both the clipping mechanism and the KL divergence term. While $\mathbb{D}_\mathrm{KL}$ represents the KL divergence term and can be further expressed as:
\begin{align}
\mathbb{D}_{\mathrm{KL}}\bigl[\pi_\theta \,\big\Vert\, \pi_{\mathrm{ref}}\bigr]
=\ &
\frac{\pi_{\mathrm{ref}}\bigl(o_{i,t}\!\mid q,\,o_{i,<t}\bigr)}%
{\pi_{\theta}\bigl(o_{i,t}\!\mid q,\,o_{i,<t}\bigr)}
\;-\; \\  &
\log\!\frac{\pi_{\mathrm{ref}}\bigl(o_{i,t}\!\mid q,\,o_{i,<t}\bigr)}%
{\pi_{\theta}\bigl(o_{i,t}\!\mid q,\,o_{i,<t}\bigr)}
\;-\;1\,. 
\end{align}
we adopt the same KL divergence term computation as in the original GRPO, but instead of using a static base model as the reference, we periodically update the reference model with the current model during training. This strategy prevents the model's updates from being constrained too closely to the base model, which could otherwise hinder performance improvements.
Policy optimization algorithms maximize the objective function $\mathcal{J}(\theta)$ via gradient ascent, which is equivalent to finding the minimum of $-\mathcal{J}(\theta)$ by gradient descent; accordingly, the RL loss function can be expressed as follows:
\begin{equation}
 \mathcal{L}_{\text{RL}} = -\mathcal{J}(\theta)
\end{equation}

\paragraph{Total loss function.} The total loss function is defined as the sum of two distinct loss components, and is computed as follows:
\begin{equation}
\mathcal{L}_{\text{Total}} = \lambda_1 \mathcal{L}_{\text{SFT}} + \lambda_2 \mathcal{L}_{\text{RL}},
\end{equation}
where $\lambda_1$ and $\lambda_2$ are two hyperparameters that balance the weights of the SFT and RL loss terms.

\paragraph{Loss Mask.} During training, the masking strategy ensures that each loss component only affects its intended segment. When computing the SFT loss, tokens outside the \textit{Guess} section are masked, enabling the language model to learn the correct labels via cross-entropy loss solely on the \textit{Guess} portion. Conversely, during RL, tokens within the \textit{Guess} section are masked in the loss calculation to prevent adverse learning signals from the guessed labels. This approach effectively isolates the two loss computations, reducing gradient conflicts during backpropagation.

\subsection{Gradient Conflict Detection and Resolution}

\begin{figure}[htbp]
	\centerline{\includegraphics[width=0.45\textwidth]{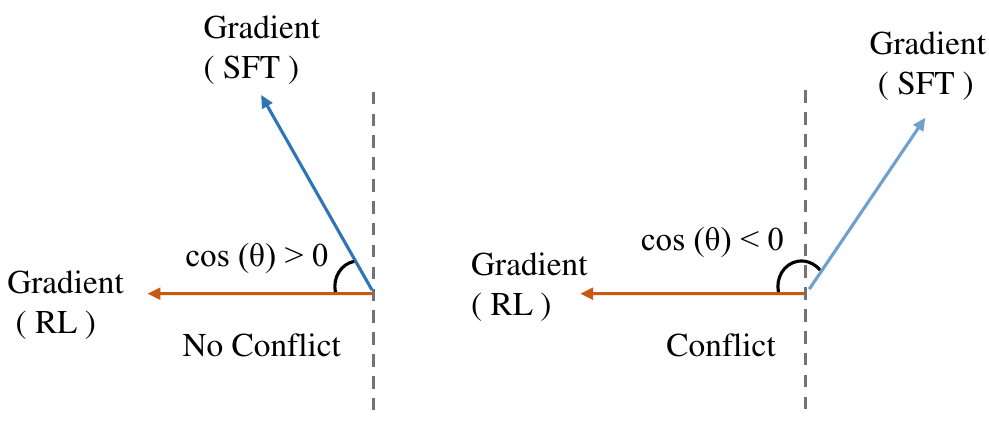}}
	\caption{Illustration of gradient conflict analysis via gradient cosine similarity}
	\label{grad}
\end{figure}

Despite the loss masking mechanism described above and the assignment of distinct weights to each objective to reduce gradient conflicts, multi-task learning cannot fully avoid such interference. To further mitigate gradient conflicts, we analyze the gradients of the two losses during backpropagation and integrate theoretical insights from PCGrad \citep{pcgrad} into the training process. As illustrated in Figure~\ref{grad}, we use cosine similarity to detect gradient conflicts: a positive cosine similarity between gradients from the two losses indicates no conflict during backpropagation, whereas a negative cosine similarity denotes the occurrence of a gradient conflict. The calculation formula of cosine similarity under the gradient can be expressed as follows:
\begin{align}
	\cos(\theta) &= \frac{\nabla \mathcal{L}_{\text{SFT}} \cdot \nabla \mathcal{L}_{\text{RL}}}{\|\nabla \mathcal{L}_{\text{SFT}}\| \cdot \|\nabla \mathcal{L}_{\text{RL}}\|}.
\end{align}
where $\theta$ is the angle between the two gradient vectors, and $\nabla$ denotes the gradient calculated by backpropagation under the corresponding loss. The final loss update rule is:
\begin{align}
	\mathcal{L}_{\text{Final}} &=
\begin{cases}
	\mathcal{L}_{\text{Total}}, & \text{\text{if }} \nabla \mathcal{L}_{\text{SFT}} \cdot \nabla \mathcal{L}_{\text{RL}} > 0,\\
	\mathcal{L}_{\text{RL}}, & \text{otherwise},
\end{cases} 
\end{align}
During parameter updates, when gradient conflicts arise, we mitigate such conflicts by retaining only the loss component associated with the RL objective.

\section{Experimental Setup}
This section outlines the experimental setup used to evaluate our approach. We describe the datasets, baseline models, and implementation details.

\subsection{Datasets}
Considering the resource constraints of our experimental process, we selected four datasets reflecting distinct classification scenarios based on recent studies \citep{chen2024open, menon2024discern, chen2025preserving}: SST-5 \citep{sst5}, Amazon \citep{amazon}, Emotion \citep{emotion}, and BBC News \citep{bbc-news}, each containing multiple categories. The detailed descriptions of these datasets are presented in Table~\ref{datas}; they cover four domains—movie reviews, intent recognition, English tweets, and News—with class counts ranging from five to eighteen.

\begin{table*}[ht]
	\centering
	\begin{tabular}{@{} l p{8.5cm} l l l @{}}
			\toprule
			\textbf{Dataset} & \textbf{Description} & \textbf{Training} & \textbf{Testing} & \textbf{Classes} \\
			\midrule
			SST-5
			& The Stanford Sentiment Treebank five-class benchmark is a standard corpus for fine-grained sentiment classification. It consists of sentences drawn from movie reviews, each annotated at the sentence level with one of five sentiment labels—very negative, negative, neutral, positive, and very positive.
			& 8,544 & 2,210   & 5   \\
			Amazon
			& This is the English (en-US) subset of the Massive Scenario Classification task from the Massive Text Embedding Benchmark (MTEB), aimed at intent prediction in voice assistant interactions. The dataset covers 18 scenario classes (such as alarm, audio, iot, calendar, play, news, and weather).
			& 11,514 & 2,974  & 18  \\
			Emotion
			& The emotion dataset is a carefully curated subset of English tweets annotated with six basic emotions—sadness, joy, love, anger, fear, and surprise—providing a standardized benchmark for evaluating emotion recognition models . Each sample consists of a tweet paired with its corresponding label.
			& 16,000 & 2,000  & 6  \\
            BBC News
            &Dataset on BBC News Topic Classification published on the BBC News website corresponding during 2004-2005. Each article is labeled under one of 5 categories: business, entertainment, politics, sport or tech.
            & 1,225 & 1,000  & 5  \\
			\bottomrule
		\end{tabular}
    	\caption{Detailed description of datasets utilized in the experimental process. Each dataset differs in terms of the number of classes, training samples, and test samples.}
	\label{datas}
\end{table*}

\subsection{Models}
Below we provide a concise summary of LLMs used in our experiments:

\noindent \textbf{Qwen2.5 (3B)\footnote{\url{https://huggingface.co/Qwen/Qwen2.5-3B-Instruct}}} is Alibaba’s open-source 3 billion parameters instruction-tuned LLM supporting long-context understanding (up to 128K tokens) and generation (up to 8K tokens), making it adept at handling extended dialogues and complex prompts. It features robust multilingual comprehension across 29 languages, ensuring broad applicability in diverse language settings. It excels in structured output generation (e.g., JSON) and instruction following.

\noindent \textbf{Qwen3 (4B)\footnote{\url{https://huggingface.co/Qwen/Qwen3-4B}}} is Alibaba’s latest open-source large language model with 4 billion parameters. Trained on a substantially larger corpus of 36 trillion tokens across 119 languages and dialects, it delivers robust performance across diverse reasoning and understanding tasks. It supports hybrid reasoning modes, seamlessly switching between CoT thinking and direct-response generation. 

\noindent \textbf{Llama3.2 (3B)\footnote{\url{https://huggingface.co/meta-llama/Llama-3.2-3B}}} is a 3 billion parameter open-source model released by Meta. It is designed as a lightweight variant of the Llama 3 family, trained on a diverse multilingual corpus and optimized for efficiency on resource-constrained environments. Despite its relatively small size, Llama3.2 demonstrates competitive performance on a wide range of reasoning and classification tasks.

\subsection{Hyperparameters}
All experiments were carried out on a multi-node cluster, each node hosting four NVIDIA L40s GPUs (48 GB each) and coordinated via DeepSpeed with ZeRO Stage 2. For SFT, we employed the open-source ModelScope Swift framework, while RL baselines used the TRL library’s GRPO implementation and our proposed GTA built atop GRPO. Inputs were truncated to a maximum of 4,096 tokens, and models were trained in bfloat16 precision with a per-device batch size of 4. In the RL phase, each prompt generated 16 candidate answers, and we applied importance sampling with a reuse factor of 4, and included a KL penalty weighted by $\beta = 0.01$ to stabilize policy updates. In our GTA, losses are assigned equal weights of 1. Each dataset is trained for 3–4 epochs.   

\section{Results and Analysis}
In this section, we report the main experimental results and provide in-depth analyses. We first compare GTA with baseline methods on multiple benchmarks, then investigate convergence behavior, reasoning robustness, and case studies to gain further insights into its effectiveness.

\begin{table*}[t]
    \centering
    \resizebox{0.68\textwidth}{!}{%
    \begin{tabular}{l|cc|cc|cc|cc}
        \toprule
        \multirow{2}{*}{Dataset} 
        & \multicolumn{2}{c}{Base} 
        & \multicolumn{2}{c}{SFT} 
        & \multicolumn{2}{c}{GRPO} 
        & \multicolumn{2}{c}{GTA} \\
        \cmidrule(lr){2-3} \cmidrule(lr){4-5} \cmidrule(lr){6-7} \cmidrule(lr){8-9} 
         & Acc & F$_1$ & Acc & F$_1$ & Acc & F$_1$ & Acc & F$_1$ \\
        \midrule
        \multicolumn{9}{l}{\textit{Qwen2.5 (3B)}} \\
        \midrule
        SST-5   & 11.76 & 13.34 & 60.72 & 59.59 & 58.60 & 57.05 & \textbf{61.58} & \textbf{61.52} \\
        Amazon  & 54.84 & 55.48 & 91.96 & 91.92 & 90.82 & 91.03 & \textbf{92.47} & \textbf{92.46} \\
        Emotion & 58.75 & 58.63 & 91.35 & 91.41 & 82.50 & 81.54 & \textbf{92.45} & \textbf{92.47} \\
        BBC News & 81.50 & 82.88 & 97.70 & 97.70 & 95.40 & 95.47 & \textbf{98.50} & \textbf{98.50} \\
        \midrule
        \multicolumn{9}{l}{\textit{Qwen3 (4B)}} \\
        \midrule
        SST-5    & 45.88 & 39.80 & 61.67 & 60.87 & 59.28 & 58.70 & \textbf{61.95} & \textbf{60.94} \\
        Amazon  & 68.96 & 70.28 & 92.57 & 92.58 & 90.55 & 90.32 & \textbf{92.87} & \textbf{92.92} \\
        Emotion  & 51.15 & 54.00 & 92.20 & 92.09 & 84.55 & 84.23 & \textbf{92.95} & \textbf{92.94} \\
        BBC News & 80.40 & 81.79 & 97.70 & 97.70 & 94.90 & 95.01 & \textbf{97.90} & \textbf{97.91} \\
        \midrule
        \multicolumn{9}{l}{\textit{Llama3.2 (3B)}} \\
        \midrule
        SST-5    & 38.42 & 36.11 & 59.91 & 50.65 & 56.33 & 53.40 & \textbf{61.18} & \textbf{60.13} \\
        Amazon   & 19.60 & 18.79 & 91.69 & 91.52 & 84.13 & 83.01 & \textbf{92.84} & \textbf{92.84} \\
        Emotion & 41.65 & 42.57 & 93.00 & 92.92 & 74.40 & 74.42 & \textbf{93.30} & \textbf{93.36} \\
        BBC News & 34.30 & 25.15 & 97.40 & 97.41 & 91.30 & 91.49 & \textbf{97.50} & \textbf{97.60} \\
        \bottomrule
    \end{tabular}%
    }
    \caption{Fine‐tuning performance (\%) on four benchmarks using SFT, GRPO, and GTA across two model sizes (Base refers to the model without fine-tuning).}
    \label{acc}
\end{table*}

\subsection{Performance}


The experimental results presented in Table~\ref{acc} evaluate the fine-tuning performance of three methods—SFT, GRPO, and GTA—across two model scales: Qwen2.5 (3B), Qwen3 (4B), and Llama3.2 (3B). The evaluation spans four classification benchmarks: SST-5, Amazon, Emotion, and BBC News. Performance metrics include accuracy and weighted F$_1$ scores. Accuracy reflects the proportion of correct predictions over the total number of instances. Weighted F$_1$ score computes the harmonic mean of precision and recall, weighted by the number of instances in each class. This metric handles class imbalance by assigning higher weights to classes with more instances.

\begin{figure}[htbp]
	\centerline{\includegraphics[width=0.5\textwidth]{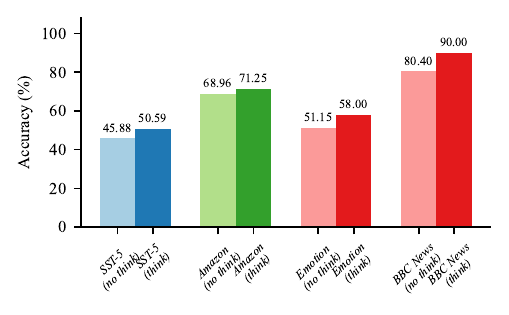}}
	\caption{Accuracy comparison of Qwen3 (4B) \textit{think} and \textit{no think} across multiple datasets}
	\label{cot}
\end{figure}

Across all datasets and different models, GTA consistently outperforms both SFT and GRPO in terms of accuracy and weighted F$_1$ Score. Notably, on the Emotion dataset, GTA achieves an F$_1$ score of 92.47\% with Qwen2.5, 92.94\% with Qwen3, and 93.36\% with Llama3.2 surpassing the other methods by a significant margin. Similarly, on the Amazon dataset, GTA attains the highest accuracy and F$_1$ scores, indicating its robustness across different domains. These results underscore the effectiveness of the proposed GTA method in enhancing model performance across diverse datasets and model scales. The consistent improvements in both accuracy and F$_1$ scores highlight GTA's robustness.

\subsection{Exploring Performance Boundaries in Zero-Shot}

We first investigated whether reasoning-enhanced prompting could improve classification performance without model fine-tuning, as this would establish important baselines and validate a core premise of our GTA framework: that explicit reasoning steps can elevate model capabilities. This exploration addresses a fundamental question in LLM deployment—whether performance limitations stem from model capabilities themselves or from suboptimal reasoning processes that can be enhanced through structured prompting. Using the base Qwen3 (4B) model with its native "think" mode toggle, we compared standard direct answering against explicit reasoning-then-answering across four benchmarks: SST-5, Amazon, Emotion, and BBC News. Figure~\ref{cot} demonstrates consistent performance improvements across all datasets when reasoning steps are incorporated. The accuracy on SST-5 increases from 45.88\% to 50.59\%, Amazon review classification improves from 68.96\% to 71.25\%, Emotion classification improves from 51.15\% to 58.00\%, and BBC News classification shows the most significant gain from 80.40\% to 90.00\%. These results reveal that models operating in a deliberate reasoning mode possess substantially higher performance ceilings than those constrained to direct response generation.

\begin{figure*}[!t]
  \centering
  \begin{subfigure}[b]{0.67\columnwidth}
    \includegraphics[width=\textwidth]{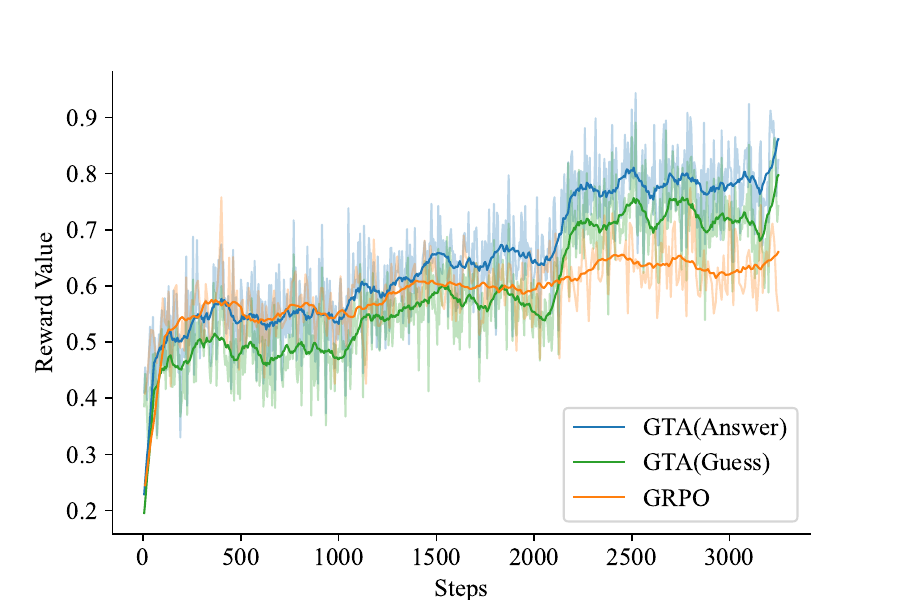}
    \caption{SST-5 (3B)}
    \label{fig:sub1}
  \end{subfigure}
  \begin{subfigure}[b]{0.67\columnwidth}
    \includegraphics[width=\textwidth]{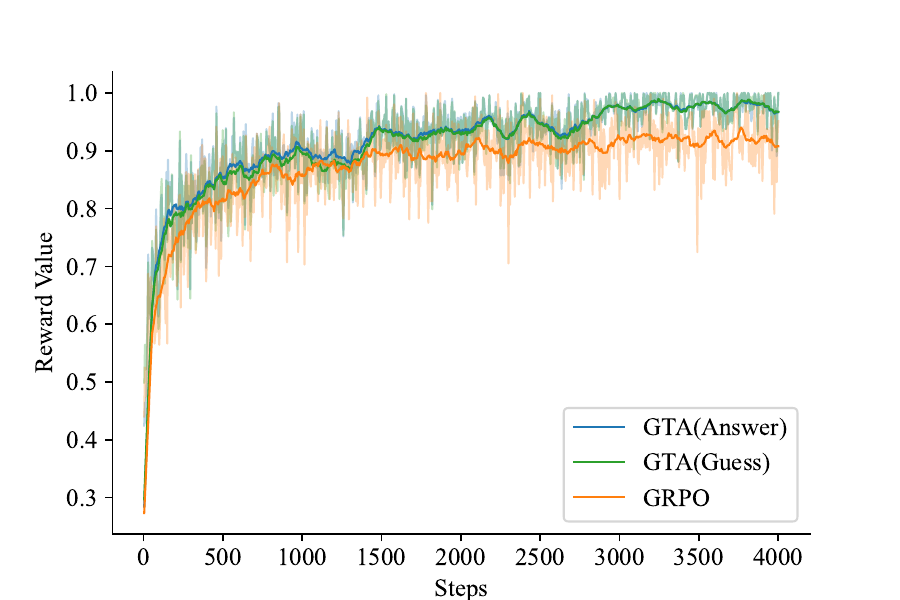}
    \caption{Amazon (3B)}
    \label{fig:sub2}
  \end{subfigure}
  \begin{subfigure}[b]{0.67\columnwidth}
    \includegraphics[width=\textwidth]{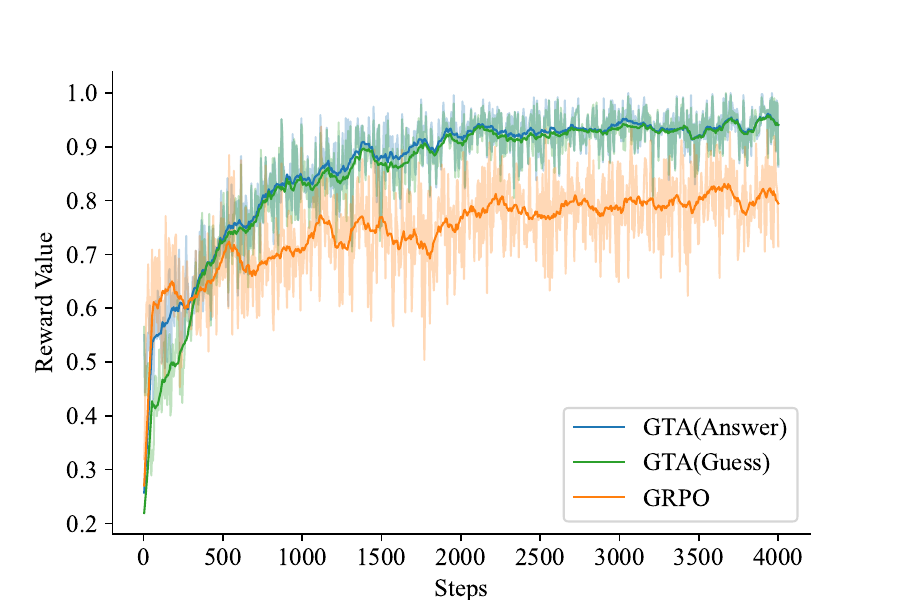}
    \caption{Emotion (3B)}
    \label{fig:sub3}
  \end{subfigure}
    \\
  \begin{subfigure}[b]{0.67\columnwidth}
    \includegraphics[width=\textwidth]{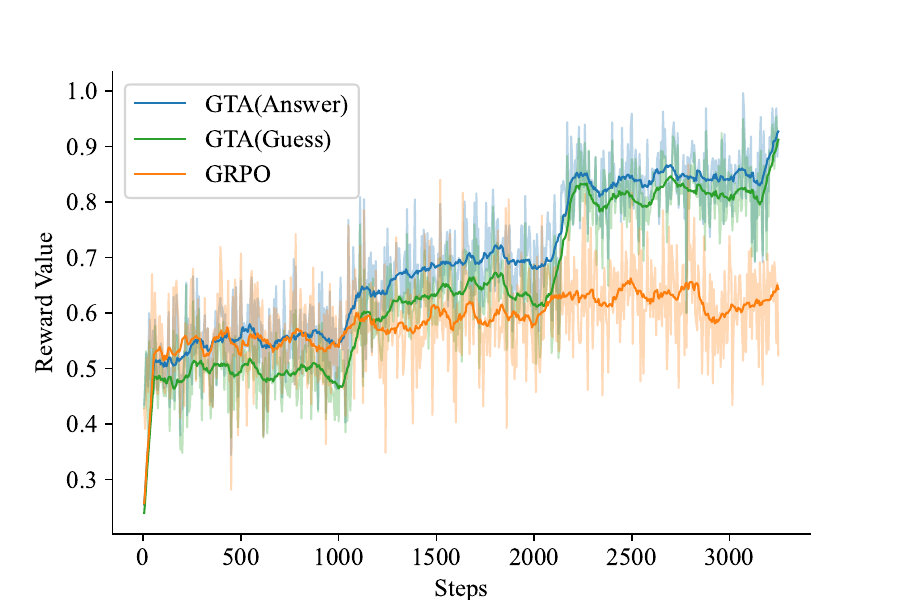}
    \caption{SST-5 (4B)}
    \label{fig:sub4}
  \end{subfigure}
  \begin{subfigure}[b]{0.67\columnwidth}
    \includegraphics[width=\textwidth]{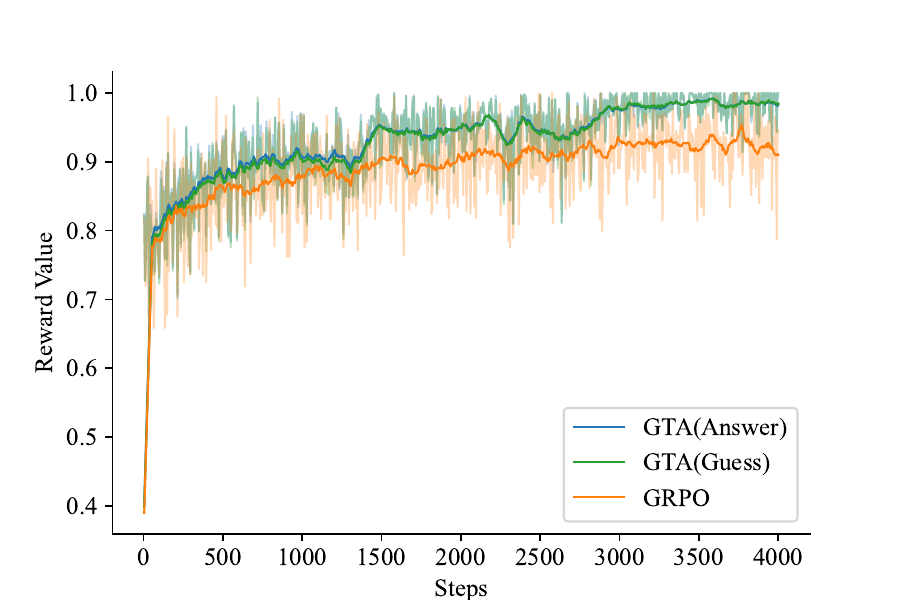}
    \caption{Amazon (4B)}
    \label{fig:sub5}
  \end{subfigure}
  \begin{subfigure}[b]{0.67\columnwidth}
    \includegraphics[width=\textwidth]{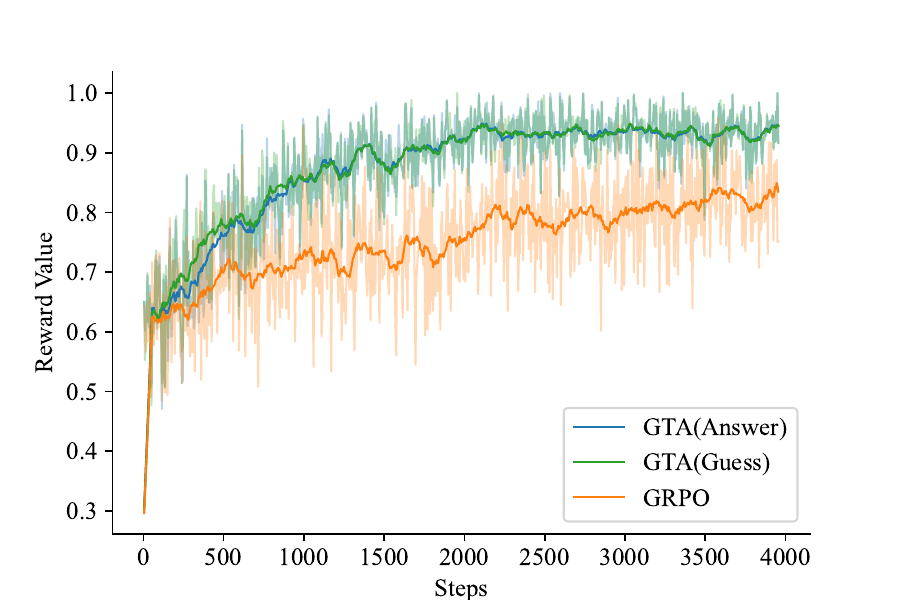}
    \caption{Emotion (4B)}
    \label{fig:sub6}
  \end{subfigure}
     \caption{Variations in accuracy reward values on the SST-5, Amazon, and Emotion datasets during GTA and GRPO fine-tuning }
  \label{con}
\end{figure*}

\begin{figure*}[!t]
	\centering
	\begin{subfigure}[b]{\columnwidth}
		\includegraphics[width=\textwidth]{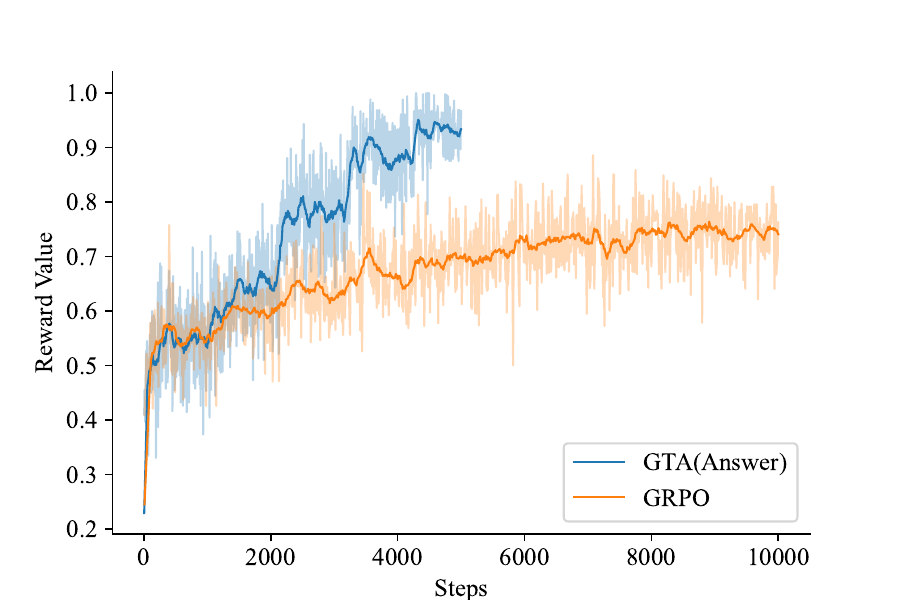}
		\caption{SST-5 (3B) Accuracy Reward}
	\end{subfigure}
	\begin{subfigure}[b]{\columnwidth}
		\includegraphics[width=\textwidth]{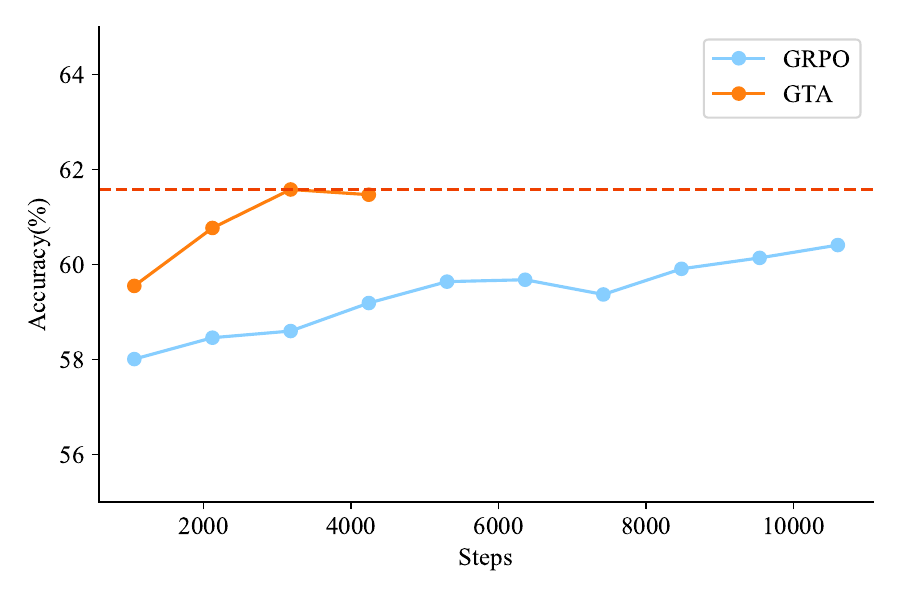}
		\caption{Accuracy (\%) on the SST-5 (3B) Test Set}
	\end{subfigure}
	\caption{Accuracy reward and test set accuracy trends of GRPO and GTA on SST-5}
	\label{analysis}
\end{figure*}

\subsection{Convergence Analysis of RL}
RL-based fine-tuning typically suffers from slow convergence due to lengthy exploration cycles. To assess how our GTA method addresses this limitation, we compared its convergence speed against the GRPO baseline across multiple datasets and model scales. Since BBC News is a relatively easy dataset with fewer categories and less training data, its convergence trends are less informative for analyzing RL training dynamics. We therefore focus our convergence analysis on SST-5, Amazon, and Emotion. Figure~\ref{con} illustrates the accuracy reward trajectories for both 3B and 4B models on SST-5, Emotion, and Amazon datasets, tracking three curves: GTA's answer segment (blue), GTA's guess segment (green), and GRPO (orange). On SST-5 and Emotion datasets, GTA demonstrates a substantial convergence advantage, with both guess and answer rewards consistently outperforming GRPO. While GRPO achieves faster initial convergence on the Amazon dataset, GTA maintains superior overall performance throughout training. These results confirm that the supervised guess segment effectively guides the answer segment, dramatically accelerating convergence while preserving the optimization benefits of RL—offering significant advantages for applying RL to language models in classification tasks.

\begin{figure*}[htbp]
	\centerline{\includegraphics[width=0.95\textwidth]{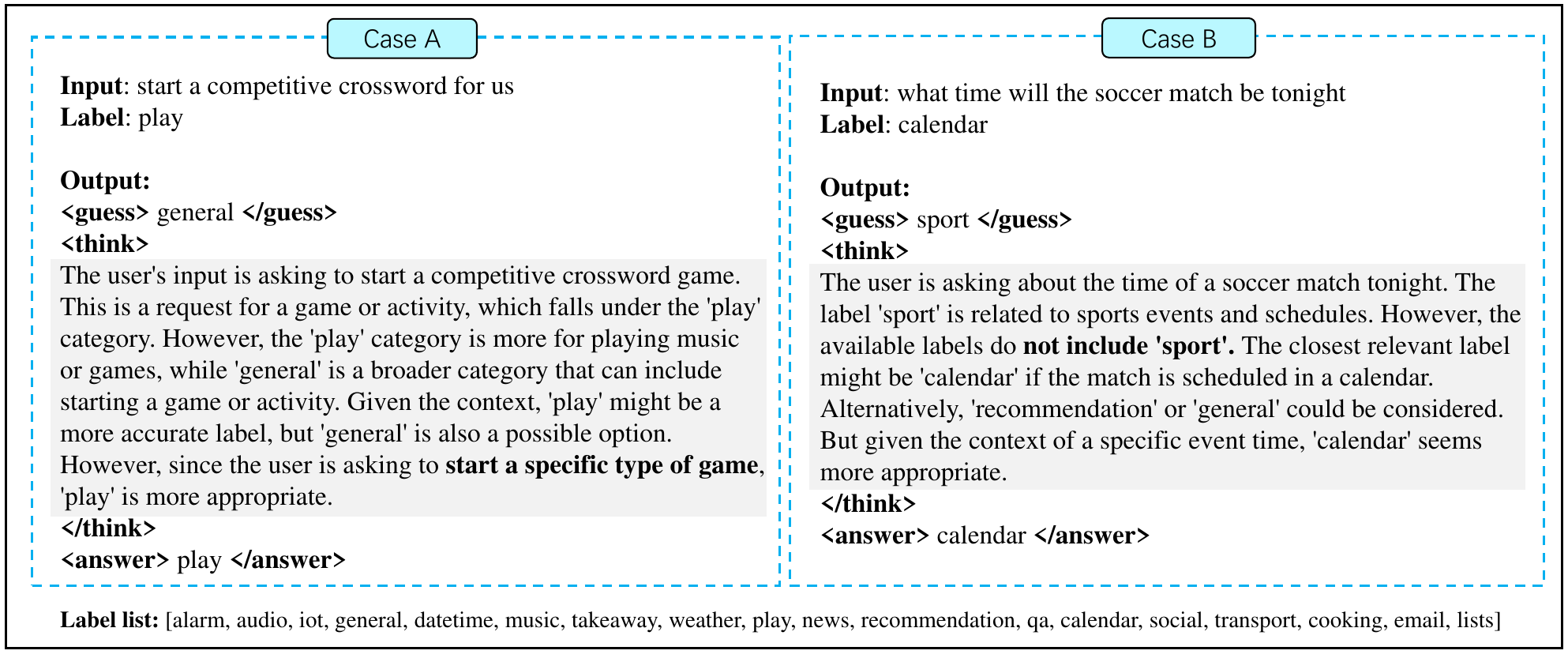}}
	\caption{Illustration of GTA reasoning: from incorrect guess to correct answer}
	\label{cot_analysis}
\end{figure*}

To investigate whether GRPO might eventually match GTA's performance given sufficient training time, we extended our experiments on the SST-5 dataset to 10,000 steps (approximately 10 epochs). Figure~\ref{analysis} presents this extended analysis using the 3B-parameter model, with Figure~\ref{analysis}(a) showing training reward progression and Figure~\ref{analysis}(b) depicting test accuracy evolution. Even in this extended training scenario, GRPO's convergence remains substantially slower than GTA's. Despite showing gradual improvements in both reward and accuracy, GRPO fails to match GTA's performance even after 10,000 steps. These findings demonstrate that GTA not only accelerates training but also achieves a higher performance ceiling, offering significant practical advantages for deploying RL in language model fine-tuning for classification tasks.
\setlength{\tabcolsep}{3.8pt}
\begin{table}
  \centering
  \begin{tabular}{cccc}
    \hline
     & GRPO&GTA (RL)&GTA (SFT+RL) \\
     \hline
    Acc & 58.60 & 57.42 & 61.58 \\
    F$_1$ & 57.05 & 56.52 & 61.52  \\       
    \hline
  \end{tabular}
  \caption{ Performance (\%)  of Qwen2.5 (3B) on GTA trained by RL on SST-5. GTA (RL) means that all stages are trained using RL, and GTA (SFT + RL) indicates that the loss of the guess is calculated through SFT.}
  \label{Ablation}
\end{table}

\subsection{Ablation on Guess}
To validate the contribution of the supervised Guess stage, we replace the \textit{Guess} component with RL updates. The results are shown in Table~\ref{Ablation}, We can observe that replacing supervised Guess with RL training does not improve the final accuracy. In fact, it yields a drop. This indicates the effectiveness of using supervised loss in the \textit{Guess} stage.


\subsection{Reasoning Process Analysis}

We find that although making a guess under the guidance of supervision can accelerate convergence, the model does not blindly commit to the guessed answer as the final prediction. As show in Figure~\ref{cot_analysis}, when the model produces an incorrect guess, subsequent reasoning steps end to allow it to correct previous mistakes. In Case A, the model first predicts an incorrect label but gradually revises its reasoning and ultimately derives the correct answer. In Case B, the model not only outputs the correct final label but also explicitly indicates that the candidate label set does not include \textit{``sport''}, thereby mitigating hallucination issues commonly observed in SFT. These examples, together with the overall accuracy improvements, highlight that GTA exhibits stronger robustness than purely SFT methods that only optimize the \textit{Guess} segment.

\section{Conclusion and Future Work}

In this work, we present GTA, a novel training framework that addresses the efficiency-capability trade-off between SFT and RL by introducing a \textit{Guess} stage to traditional CoT outputs. Under this framework, model outputs are organized into \textit{Guess}, \textit{Think}, and \textit{Answer} segments, where the \textit{Guess} is optimized via SFT while the overall format and final output are optimized through RL. To mitigate gradient conflicts, we apply loss masking and cosine-similarity constraints. Experimental results on four text-classification benchmarks show that GTA consistently outperforms pure SFT and GRPO baselines in accuracy and F$_1$ scores while achieving significantly faster convergence than pure RL approaches. Through analysis of training dynamics, we demonstrate that GTA successfully combines the efficiency of supervised learning with the performance gains of RL, substantially accelerating convergence and addressing the exploration inefficiency inherent to RL-based fine-tuning. Theoretically, our approach is not limited to text classification tasks, and we plan to explore extending GTA to broader NLP tasks in future work.

\section*{Limitations}
While publicly available text classification datasets often contain noisy data, we did not perform preprocessing or sample high-quality subsets in this study. Additionally, due to resource constraints, we validated our proposed method only on 3B and 4B parameter models, without extending the evaluation to larger-scale models. Given that RL heavily relies on the underlying model's capacity, more powerful and generalizable models may yield greater benefits.
\bibliography{custom}

\appendix


\end{document}